\documentclass[a4paper, 10 pt, conference]{ieeeconf}  

\usepackage{soul}
\usepackage{siunitx}
\usepackage{tabularx, booktabs}
\usepackage{multirow}
\usepackage{rotating}
\usepackage{graphicx}
\usepackage{gensymb}
\usepackage[caption=false, font=footnotesize]{subfig}
\usepackage[para,online,flushleft]{threeparttable}
\usepackage{lipsum}

\usepackage{todonotes}

\IEEEoverridecommandlockouts                              

\title{\LARGE \bf
BirdNet+: End-to-End 3D Object Detection in LiDAR Bird's Eye View
}

\author{Alejandro Barrera$^{1}$, Carlos Guindel$^{1}$, Jorge Beltr\'an$^{1}$ and Fernando Garc\'i{}a$^{1}$ \thanks{$^{1}$Authors are with the Intelligent Systems Lab (LSI) research group, Universidad Carlos III de Madrid, Spain {\tt\small alebarre@pa.uc3m.es, \{cguindel,jbeltran,fegarcia\}@ing.uc3m.es}} }

\begin{document}

\maketitle
\thispagestyle{empty}
\pagestyle{empty}

\begin{abstract}
On-board 3D object detection in autonomous vehicles often relies on geometry information captured by LiDAR devices. Albeit image features are typically preferred for detection, numerous approaches take only spatial data as input. Exploiting this information in inference usually involves the use of compact representations such as the Bird's Eye View (BEV) projection, which entails a loss of information and thus hinders the joint inference of all the parameters of the objects' 3D boxes. In this paper, we present a fully end-to-end 3D object detection framework that can infer oriented 3D boxes solely from BEV images by using a two-stage object detector and ad-hoc regression branches, eliminating the need for a post-processing stage. The method outperforms its predecessor (BirdNet) by a large margin and obtains state-of-the-art results on the KITTI 3D Object Detection Benchmark for all the categories in evaluation. 


\end{abstract}

\section{INTRODUCTION}
3D object detection has been gaining popularity recently in the field of on-board perception, commensurately with the high relevance of the task for autonomous driving. Whereas former object detection approaches were frequently restricted to the 2D image coordinates, modern methods go far beyond and try to provide an accurate estimation of the location and dimensions of the objects in the environment, which are usually represented as 3D cuboids. This information is critical to enable safe and reliable navigation in all kinds of traffic situations, including crowded environments with a multitude of other road users.

Among the different sensor modalities used in automotive applications, LiDAR stands as an ideal candidate for this task due to its high measurement accuracy and robustness. Although less bulky than visual information, the processing of LiDAR data is nevertheless not without its challenges. Representations such as the Bird's Eye View (BEV) allow the efficient use of deep learning inference frameworks for feature extraction and inference. 

\begin{figure}[tb]
\centering
\includegraphics[width=0.99\linewidth]{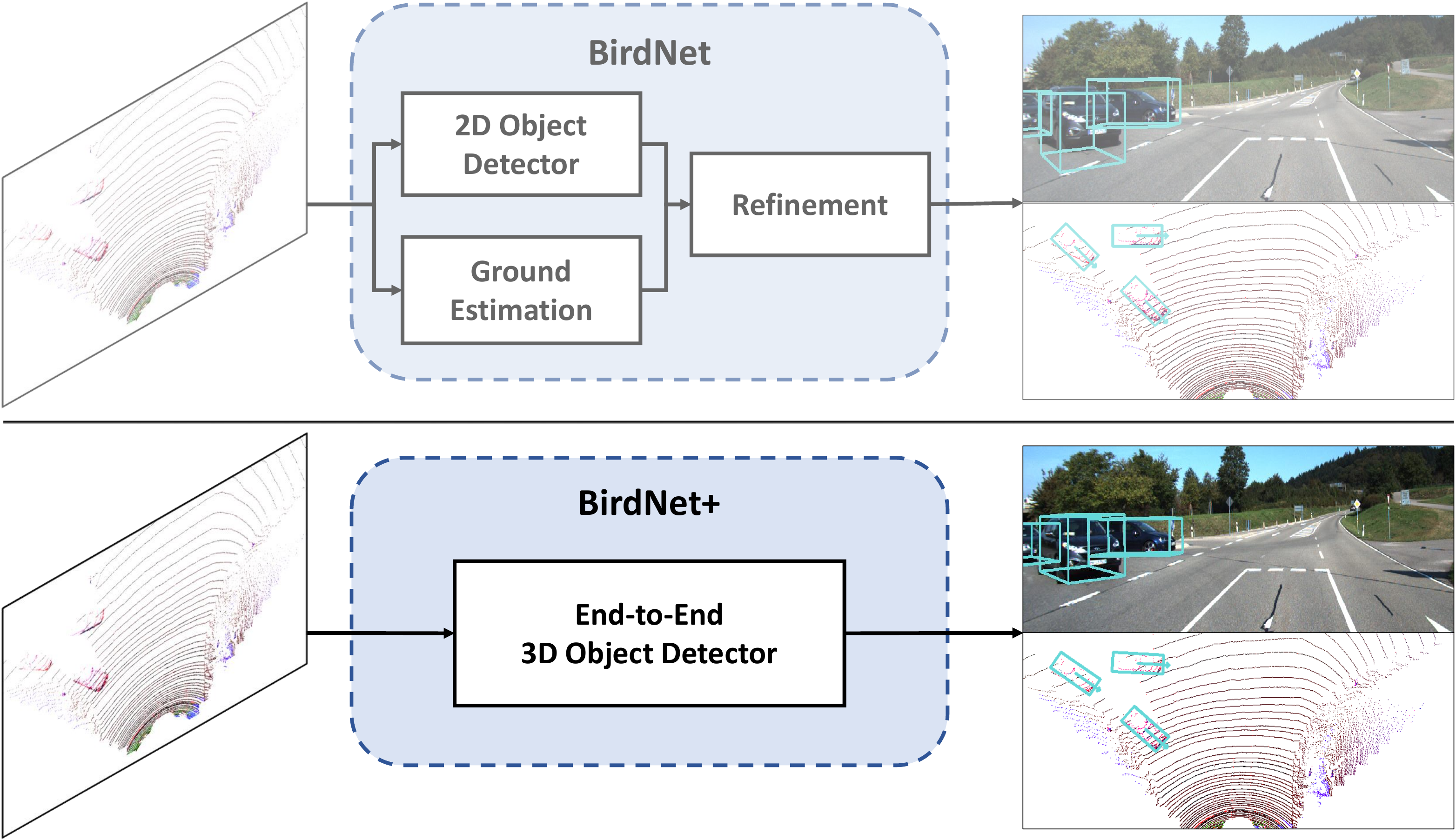}
\caption{Previous framework vs. proposed approach for 3D detection}
\label{fig:differences}

\end{figure}

In this context, BirdNet was introduced in \cite{Beltran2018} as an object detection framework aimed to provide 3D detections using BEV data only. The method proved the adequacy of applying an image-based detector (Faster R-CNN \cite{Ren2015a}) to the processing of BEV structures, although it also had some limitations stemming from its design. Hence, despite being intended to work mostly on an end-to-end basis, BirdNet still required a hand-crafted post-processing step to obtain the final rotated 3D box representing each obstacle, as shown in Fig.~\ref{fig:differences}.

This work introduces a set of proposals aimed to enhance the original BirdNet approach, with the final goal of building a truly end-to-end approach for 3D object detection on LiDAR data. In particular, we update some of the building blocks of the original proposal following the current trends in the literature and modify its architecture to embed the former post-processing stage into the inference procedure, thus improving the detection performance significantly. 

The main contributions of this paper are:
\begin{itemize}
    \item We introduce a novel strategy to perform regression of an oriented box using a two-stage detection approach. Thus, while the proposals generated in the first stage are axis-aligned, features pooled from them are used in the second stage to estimate the parameters of the final rotated box. 
    \item Although the BEV representation is generally problematic in grasping the features related to the height coordinate, we show the feasibility of training an end-to-end pipeline able to extract from it all the parameters defining the objects' bounding boxes, including their height and elevation. In this way, further post-processing steps can be avoided. 
\end{itemize}

As usual, we rely on the challenging KITTI 3D Object Detection Benchmark \cite{Geiger2012KITTI} to assess the adequacy of the adopted solutions.

\section{RELATED WORK} \label{related_works}

LiDAR data provide an accurate representation of the scene, enabling a full understanding of the traffic situation. However, the size and sparsity of the captured point cloud make it difficult to be processed efficiently. As a consequence, state-of-the-art 3D object detection methods based on LiDAR information have opted for different kinds of input formats, using both raw and grid-like representations of the laser cloud.



Methods that process the raw cloud are able to exploit the detailed geometry and reflection data to classify and estimate the 3D pose of the agents in the scene using PointNet-like networks. However, the large amount of points captured by modern LiDAR devices leads to high computational costs. In order to cope with this burden, some works reduce the cloud size by isolating regions of interest previously extracted with the help of 2D image detections \cite{Qi2017a} or RGB semantic segmentation \cite{Yang2018b}. Point R-CNN \cite{Shi2019} performs a point-wise segmentation to remove the background before estimating 3D boxes. Recently, STD \cite{Yang2019a} combined a light per-point feature extraction stage over the whole cloud with a second voxelization step that alleviates the computation load while providing excellent performance.

To further speed up the object detection task, a different group of approaches \cite{Zhou2017a, Yan2018} takes a voxelized LiDAR cloud as input. This procedure creates a volumetric grid, reducing both the size and the sparsity of the data and facilitating their processing. On the contrary, the subdivision of the 3D space in equally-sized regions entails a loss of information and generates many empty voxels at far distances. These approaches take advantage of the structuredness of the input by applying 3D convolutions to features extracted via voxel feature encoders. However, despite the cloud discretization, the use of three-dimensional inputs still demands a significant computational load.

In order to tackle the limitations related to the high computational costs, PointPillars \cite{Lang2018} uses a simplified PointNet to produce a 2D feature map from 3D pillars (voxels of infinite height), which is then fed into a single-stage object detector. In this fashion, some works use the Bird's Eye View (BEV) projection of the LiDAR data with a hand-crafted encoding to feed either single- \cite{Yan2018, Simon2018} or two-stage \cite{Beltran2018, Wirges2018} image detectors. MODet \cite{Zhang2019} pushes the limits of this trend using an even more compressed (binary) representation of the BEV. These structures reduce the sparsity of data and are simpler to process, which enables their use in real-time applications; however, some information, especially that related to the height coordinate, is inevitably lost in the simplification. 
Thus, although these approaches obtain promising detection results in the Bird's Eye View projection, they usually find difficulties in the estimation of final 3D boxes, whose height and elevation are often computed at a later stage.

\section{PROPOSED APPROACH}\label{proposed_approach}
\begin{figure*}[tb]
\centering
\includegraphics[width=\linewidth]{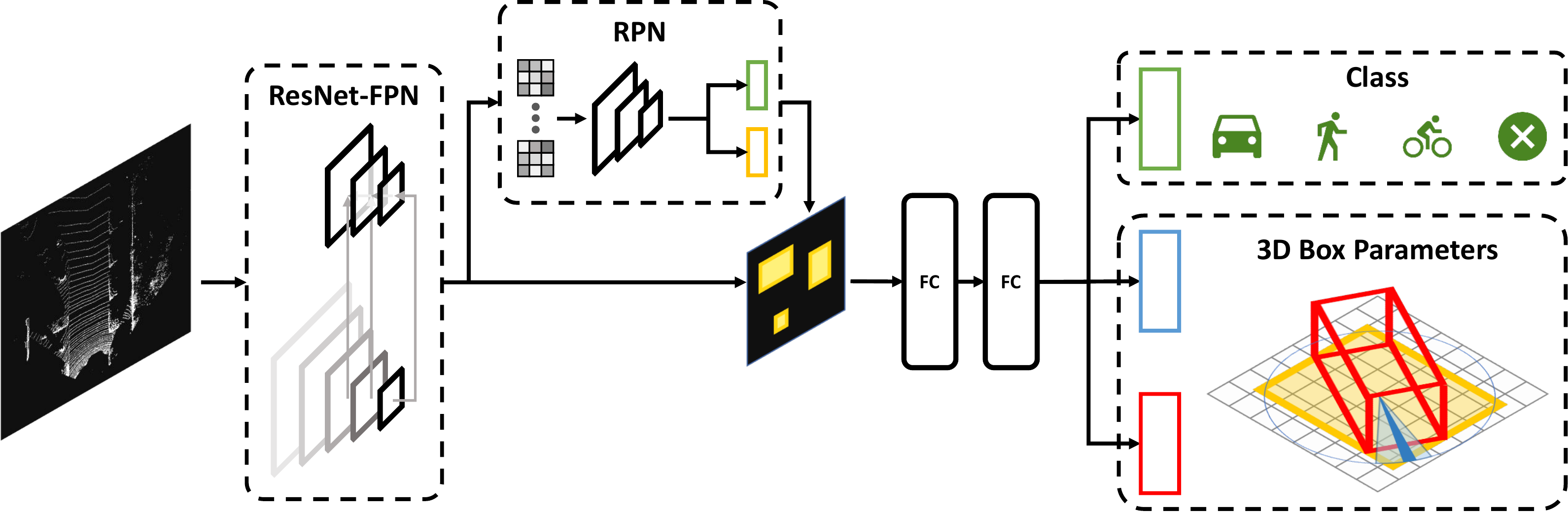}
\caption{Overview of the proposed approach}
\label{fig:network_diagram}
\end{figure*}

The method proposed in this paper, depicted in Fig.~\ref{fig:network_diagram}, uses Bird's Eye View (BEV) images as input, and is able to perform 3D object detection in a two-stage procedure. In the first stage, a set of axis-aligned proposals is obtained. Afterward, the second stage estimates the classification and rotated 3D box (i.e., centroid, dimensions, and orientation) corresponding to each proposal.


\subsection{Bird Eye’s View Representation} \label{bev_representation}
The procedure to generate the BEV images derives from the one presented in \cite{Beltran2018}. This representation codifies the LIDAR point cloud into 2D structures with three different channels: maximum height (up to \SI{3}{\meter} from the theoretical ground plane), mean intensity, and normalized density of the points within the cell.  It should be noted that this encoding does not provide information about the lowest point in each cell, i.e., the position of the ground.


In the selected configuration, space is divided into a grid of square cells representing \SI{5}{\centi\meter} each. In our experiments on the KITTI dataset, we limit the range to \SI{35}{\meter} in every direction except backwards, and keep only those points in the field of view of the camera, where annotations are available. 



\subsection{Inference Framework} \label{inference_framework}
As in the original work, the proposed approach uses the Faster R-CNN meta-architecture \cite{Ren2015a} to perform detection on BEV data. Faster R-CNN is designed to accept RGB images as input and, therefore, is naturally suited to handle the 2D BEV structures. As said before, Faster R-CNN is composed of two stages aimed to generate proposals and to classify and refine them, respectively. Both of them rely on a shared set of features extracted from the input data. 

In this section, the modifications introduced at the different stages of the pipeline to enable end-to-end 3D box estimation will be presented. It should be noted that, although we will treat them as incremental steps, they are not mutually dependent and could be implemented separately.


    
    \medskip
    
    \noindent\textbf{Feature extraction.} In Faster R-CNN, input data is first fed to a feature extractor responsible for computing the feature maps on which inference will be based. We replace the VGG-16 featured by the original BirdNet with a ResNet-50 \cite{He2015b} that offers a better tradeoff between accuracy and computation speed and is less prone to overfitting than other alternatives. Typically, features are extracted from ResNet-50 at the \textit{conv4} layer, where they have been downsampled by a factor of $16$, thus being unsuitable for the detection of small obstacles, such as pedestrians, in BEV maps. Two different alternatives are proposed to overcome this limitation:
    \begin{itemize}
        \item Retrieving feature maps from the \textit{conv3} layer, where the downsampling factor is $8$, as in the original BirdNet. Although this solution improves the resolution, it has proven insufficient for pedestrian detection. 
        \item Taking advantage of a Feature Pyramid Network (FPN) \cite{Lin2016} so that features corresponding to each object are extracted from every ResNet block output until C4, which allows combining features at different scales. 
    \end{itemize}

    \medskip
    
    \noindent\textbf{Region proposal.} In the first stage, the Region Proposal Network (RPN) generates proposals from a set of predefined anchors. Anchor sizes have been selected according to \cite{Beltran2018}, featuring three scales ($16^2$, $48^2$ and $80^2$) and three aspect ratios (1:1, 1:2, 2:1). These anchors are axis-aligned boxes, and so are the proposals provided by the RPN. For feature pooling, we adopt ROIAlignV2 \cite{wu2019detectron2}, which offers a slightly better image-feature-map alignment than the previous version. A $7 \times 7$ pooling resolution is employed. 
    
    
    \medskip
    
    
    \noindent\textbf{Classification and bounding box regression.} The RPN at the first stage provides proposals represented by 2D bounding boxes in BEV coordinates. The second stage is responsible for classifying these proposals and, most notably, regress a 3D box representing the object. We let this prediction step be composed of two Fully Connected (FC) layers, with 1,024 elements each, whose output is ultimately fed to a set of individual heads. Each of these heads is made of an FC layer and is in charge of a different task. The bare-bones model, which follows the original BirdNet, includes three branches for classification, axis-aligned box regression, and discretized yaw classification, respectively. 
        
    
    \medskip
    
    \noindent\textbf{Real dimensions regression.} As a first step towards the complete removal of the post-processing stage, we modify the axis-aligned box regression branch to predict the dimensions of the rotated box enclosing each object instead of the axis-aligned box employed by the RPN. This approach avoids the redundancy of the per-vertice regression used by other methods \cite{Chen2016b, Ku2018} by estimating only the center coordinates and dimensions of the boxes. 
    
    Therefore, we let the regression branch estimate the ($x,y,l,w$) parameters of each object, with $l$ being its length and $w$ its width in BEV units (i.e., scaled by the grid resolution). Identical to the original box estimation branch, the regression targets are defined as offsets from the RPN proposals. Along with the yaw estimation provided by the corresponding branch, these parameters fully define the rotated bounding box representing the object in the BEV map. Remarkably, the feature vector used for this regression task is obtained via ROIAlign by pooling features from an axis-aligned proposal; therefore, this branch effectively learns the refinement step between axis-aligned boxes and object-aligned detections that was formerly performed at post-processing.
    
    In line with the new nature of detections, the final non-maximum-suppression (NMS) stage over the set of resulting detections has been replaced by a rotated version that works on a per-category basis. The IoU threshold is set to 0.3.

    \medskip
    
    \noindent\textbf{Height and vertical position regression. } 
    The last two parameters that must be embedded into the inference framework to dispense with the post-processing stage completely are the height and vertical position of the centroid of the object's 3D box. We have included a new prediction task to perform regression of these parameters. Due to the BEV encoding in use, which lacks ground information and only considers the height of the taller object in the cell, this is a particularly challenging task. However, we have proven the effectiveness of the inference framework to provide a reasonably 3D box estimate under these constraints.
    
    The approach to this task mimics the original box regression branch; thus, the regression targets, $\Delta h$ and $\Delta z$, are the differences in height ($h$) and vertical position ($z$) of the centroid of the object's cuboid with respect to a reference box, as shown in (\ref{eq:delta1}) and (\ref{eq:delta2}).
    \begin{equation}
        \Delta h = w_{h} \cdot \frac{\ln{(h)}}{h_\text{ref}}
        \label{eq:delta2}
    \end{equation}
    \begin{equation}
        \Delta z = w_{z} \cdot \frac{z - z_\text{ref}}{h_\text{ref}}
        \label{eq:delta1}
    \end{equation}

    Following \cite{Ren2015a}, weights $w_{z}$ and $w_{h}$ aim to normalize the regression targets so that they have a variance close to $1$. Note that these targets are identical to the ones used to estimate the size and position of the BEV detections.
    
    The reference box is assigned a height $h_\text{ref}$ equal to the average height of the training samples belonging to that category and positioned ($z_\text{ref}$) lying on the theoretical ground plane. The average values of height in the KITTI dataset are \SIlist{1.53; 1.76; 1.74}{\meter} for Car, Pedestrian, and Cyclist, respectively.

    \medskip
    
    \noindent\textbf{Hybrid yaw angle estimation.} The estimation of the yaw angle is a requirement to obtain not only the rotated boxes representing the obstacles in the BEV image but also the final 3D boxes. The baseline BirdNet \cite{Beltran2018} was endowed with a yaw estimation branch that performed category-aware multinomial classification over a set of bins resulting from the discretization of the 360\degree\ range of possible values. This approach showed good robustness but suffered from a limited resolution that depended on the number of bins.

    We preserve this bin classification task but also add a complementary branch for the regression of the residual error between the center of the predicted bin and the ground-truth yaw angle. This additional step allows closing the gap between the coarse prediction of the yaw classification branch and the actual orientation of the object.
    
    
    As in the original approach, bins are selected so that their centers straightforwardly represent the cardinal orientations (forward/rear and left/right). Regarding the regression task, predictions are both category-aware and bin-aware; i.e., a residual is estimated for each bin and each category. These residuals are normalized to the unit. In our experiments, we use 12 bins, down from the 16 featured by the original BirdNet. Thus, the classification problem is simplified whereas final accuracy increases thanks to the regression estimation.
    

    

\subsection{Multi-Task Training}
We apply a multi-task loss to train all the network tasks together, as shown in (\ref{eq:loss}).
\begin{equation} \label{eq:loss}
    L = L_\text{rpn} + L_\text{cls} + L_\text{bbox} + L_\text{yaw} 
\end{equation}


In this equation, $L_\text{rpn}$ accounts for both anchor classification and bounding box regression at the RPN. $L_\text{cls}$ is the multiclass classification loss among all the available classes, including the background. $L_\text{bbox}$ includes the regression of the six parameters defining the 3D bounding box: the dimensions $l$, $w$, and $h$ of the object and the $(x,y,z)$ coordinates of its centroid. As introduced in the previous section, offsets for the regression of $(l,w,x,y)$ are computed from the axis-aligned RPN proposals, whereas $(h,z)$ offsets are relative to a reference 3D box with average dimensions. Finally, $L_\text{yaw}$ is composed of two components for bin classification term and residual regression, respectively. All the classification tasks use unweighted cross-entropy losses, while regression tasks rely on smooth-L1 losses normalized over the total number of regions.

Following \cite{Beltran2018}, we fine-tune the models from an ImageNet \cite{imagenet_cvpr09} pre-trained model. For the new weights, we use the Xavier initialization with a normal distribution \cite{glorot2010understanding}. As usual, the training set has been augmented by random horizontal flipping, doubling the set of samples.

\section{Experimental Results} \label{experiments}
The proposed approach has been validated through a set of experiments carried out on the KITTI Object Detection Benchmark \cite{Geiger2012KITTI}, focusing on the 3D and Bird's Eye View detection tasks. We use an implementation based on Detectron 2 \cite{wu2019detectron2} to take advantage of the different optimizations included in that framework. The analysis is divided into two parts: first, we evaluate the effectiveness of each of the solutions introduced in Sec.~\ref{inference_framework}; and then, the approach is compared with other methods in the KITTI leaderboard.

\subsection{Ablation Studies}
In this section, we aim to investigate the effect of the modifications over the baseline framework separately. We adopt the KITTI training set for both training and validation following the split in \cite{Chen2016b}. Note that we follow the KITTI evaluation criteria regarding IoU overlapping, so that car detections are required a minimum overlap of 0.7. All models were trained by Stochastic Gradient Descent (SGD) for $40,000$ iterations, with a batch size of $4$ and a learning rate of $0.01$.

Table~\ref{tab:paramselection} shows the impact of the incremental improvements introduced in this paper, in terms of both BEV and 3D detection performance. Note that every model also includes the features of the previous ones. The baseline results from BirdNet were obtained again and differ from those in the paper \cite{Beltran2018} as we fixed some issues related to the projection of the results in the camera frame.

Introducing ResNet-50 leads to a very significant increase in the Average Precision (AP) stats for both pedestrians and cyclists, whose representation in BEV is smaller. Further improvements are achieved with the FPN variant, as these categories benefit from the enhanced resolution provided by the multiscale feature pooling. On the other hand, the introduction of the real dimensions regression yields a significant performance boost in 3D detection. This enhancement, which is increased with the introduction of height regression, confirms the suitability of the end-to-end approach for 3D box estimation proposed in this work. Finally, the yaw refinement via regression of the residual proves to be a useful alternative to refine the quality of the detections. 




%


\begin{table}[htb] 
	\caption{BEV and 3D detection performance (AP BEV \% and AP 3D \%) on the KITTI validation set for different variants.}
	\label{tab:paramselection}
	\centering
	\begin{tabular}{l c c c c c c }
		\toprule
		& \multicolumn{2}{c}{Car} & \multicolumn{2}{c}{Pedestrian} & \multicolumn{2}{c}{Cyclist} \\  
		\cmidrule(lr){2-3} \cmidrule(l){4-5} \cmidrule(l){6-7}
		 & BEV & 3D & BEV & 3D & BEV & 3D \\ 
		 \midrule   
		 BirdNet \cite{Beltran2018} & 63.6 & 37.1 & 42.8 & 35.3 & 34.2 & 31.8 \\ 
         +ResNet-50 & 64.6 & 34.6 & 55.8 & 46.5 & 40.3 & 36.0  \\ 
         +FPN & 63.0 & 35.4 & 58.4 & 48.0 & 42.7 & 38.4 \\
         +real dimensions & 64.4 & 40.6 & 62.2 & 51.2 & 44.0 & 37.6 \\
         +height & 63.8 & 51.3 & \textbf{62.8} & \textbf{52.7} & 41.9 & 40.0 \\
         +yaw residual & \textbf{66.3} & \textbf{55.6} & 61.7 & 52.4 & \textbf{44.9} & \textbf{42.6} \\
		\bottomrule
	\end{tabular}
\end{table}

\subsection{Overall Assessment}

Table~\ref{tab:bigcomparison} shows the performance of the final configuration, with all the features, on the KITTI benchmark (testing set), along with the results obtained by comparable methods. We limit the analysis to methods based on the use of the Bird's Eye View projection. 

\begin{table*}[htb]
	\caption{Evaluation results for 3D localization and 3D detection performance (AP BEV and AP 3D) on the KITTI benchmark (test set)}
	\label{tab:bigcomparison}
	\centering
    \begin{threeparttable}
	\begin{tabular}{l l l ccc ccc r}
		\toprule
		& & & \multicolumn{3}{c}{AP 3D (\%)} & \multicolumn{3}{c}{AP BEV (\%)} \\  
		\cmidrule(lr){4-6} \cmidrule(lr){7-9} 
		Cat. & Method & Input data  & Easy & Mod. & Hard & Easy & Mod. & Hard & T (ms) \\ 
		\midrule   
		\multirow{8}[3]{*}{Car} & MODet \cite{Zhang2019} & LiDAR (BEV)  & - & - & - & 90.80 & 87.56 & 82.69 & 50 \\
		& PIXOR++ \cite{Yang2018} & LiDAR (BEV) & - & - & - & 93.28 & 86.01 & 80.11 & 35 \\
	    & AVOD-FPN \cite{Ku2018}\tnote{*} & RGB + LiDAR & 83.07 & 71.76 & 65.73 & 90.99 & 84.82 & 79.62 & 100 \\
	    & MV3D (L) \cite{Chen2016b}& LiDAR (BEV+FV) & 68.35 & 54.54 & 49.16 & 86.49 & 78.98 & 72.23 & 240 \\
		& C-YOLO \cite{Simon2019} & RGB + LiDAR & 55.93 & 47.34 & 42.60 & 77.24 & 68.96 & 64.95 & 60 \\
		& TopNet-Ret. \cite{Wirges2018} & LiDAR (BEV) & - & - & - & 80.16 & 68.16 & 63.43 & 52 \\
		\cmidrule(lr){2-10}
		& BirdNet \cite{Beltran2018} & LiDAR (BEV) & 40.99 & 27.26 & 25.32 & 84.17 & 59.83 & 57.35 & 110 \\
		& BirdNet+ (ours) & LiDAR (BEV) & 70.14 & 51.85 & 50.03 & 84.80 & 63.33 & 61.23 & 100 \\	
		\midrule
        \multirow{5}[3]{*}{Ped.} & AVOD-FPN \cite{Ku2018}\tnote{*} & RGB + LiDAR & 50.46 & 42.27 & 39.04 & 58.49 & 50.32 & 46.98 & 100 \\
        & C-YOLO \cite{Simon2019} & RGB + LiDAR & 17.60 & 13.96 & 12.70 & 21.42 & 18.26 & 17.06 & 60 \\ 
        & TopNet-Ret. \cite{Wirges2018} & LiDAR (BEV) & - & - & - & 18.04 & 14.57 & 12.48 & 52 \\
        \cmidrule(lr){2-10}
		& BirdNet \cite{Beltran2018} & LiDAR (BEV) & 22.04 & 17.08 & 15.82 & 28.20 & 23.06 & 21.65 & 110 \\
		& BirdNet+ (ours) & LiDAR (BEV) & 37.99 & 31.46 & 29.46 & 45.53 & 38.28 & 35.37 & 100 \\
		\midrule 
        \multirow{5}[3]{*}{Cyc.} & AVOD-FPN \cite{Ku2018}\tnote{*} & RGB + LiDAR & 63.76 & 50.55 & 44.93 & 69.39 & 57.12 & 51.09 & 100 \\
        & TopNet-Ret. \cite{Wirges2018} & LiDAR (BEV) & - & - & - & 47.48 & 36.83 & 33.58 & 52 \\
        & C-YOLO \cite{Simon2019} & RGB + LiDAR & 24.27 & 18.53 & 17.31 & 32.00 & 25.43 & 22.88 & 60 \\ 
        \cmidrule(lr){2-10}
		& BirdNet \cite{Beltran2018} & LiDAR (BEV) & 43.98 & 30.25 & 27.21 & 58.64 & 41.56 & 36.94 & 110 \\
		& BirdNet+ (ours) & LiDAR (BEV) & 67.38 & 47.72 & 42.89 & 72.45 & 52.15 & 46.57 & 100 \\	
		\bottomrule
	\end{tabular}
    \begin{tablenotes}
		\item[*] AVOD makes use of two separate models: one for Car and another for Pedestrian and Cyclist detection.
	\end{tablenotes}
    \end{threeparttable}
    \vspace{-1em}
\end{table*}

The 3D and BEV detection results obtained by the proposed approach are comparable to the ones provided by the other state-of-the-art methods. However, the proposed framework is the only one among those presented here that can estimate 3D boxes for all the evaluated categories using only LiDAR BEV images. Unlike other methods, our framework does not rely on additional sources of data (e.g., images) and is designed to perform multiclass detection using a single model and just one BEV representation (with fixed grid resolution).

As shown in the table, runtime per frame for the final configuration (including FPN) is around 100 ms on an NVIDIA Titan Xp, which proves the suitability of the adopted two-stage detection for online processing. Additionally, as a consequence of this design, our method presents notable results in the detection of agents whose BEV projection becomes small, i.e., pedestrians and cyclists.

\begin{figure*}
\subfloat{
\\
\includegraphics[width=0.328\linewidth]{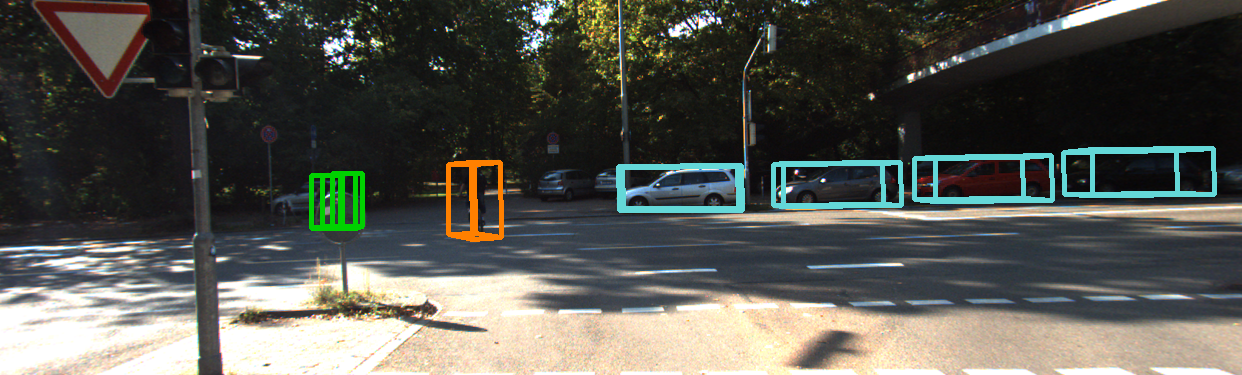}
\includegraphics[width=0.328\linewidth]{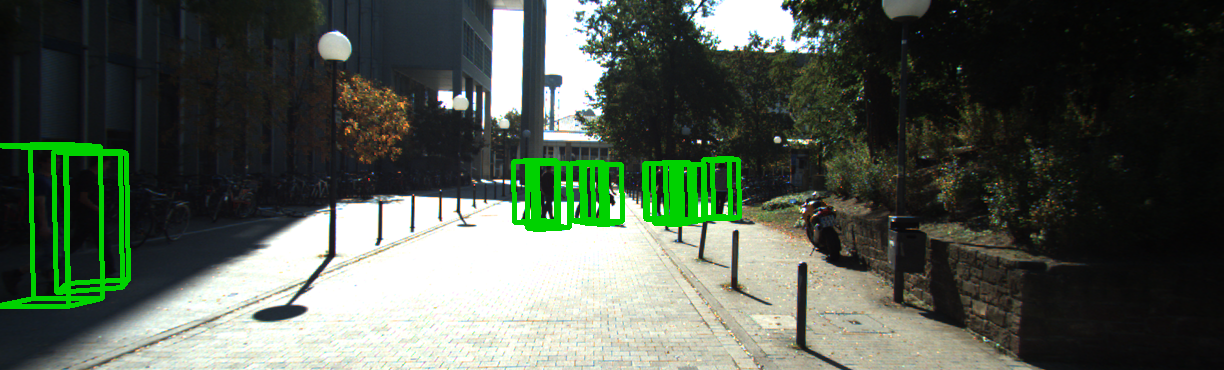}
\includegraphics[width=0.328\linewidth]{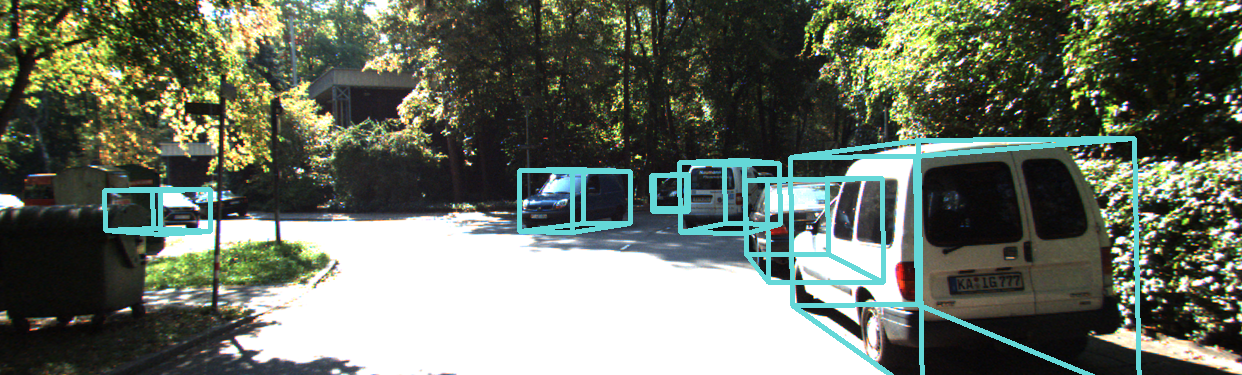}
}
\setcounter{subfigure}{0}
\\
\subfloat[]{%
\includegraphics[width=0.328\linewidth]{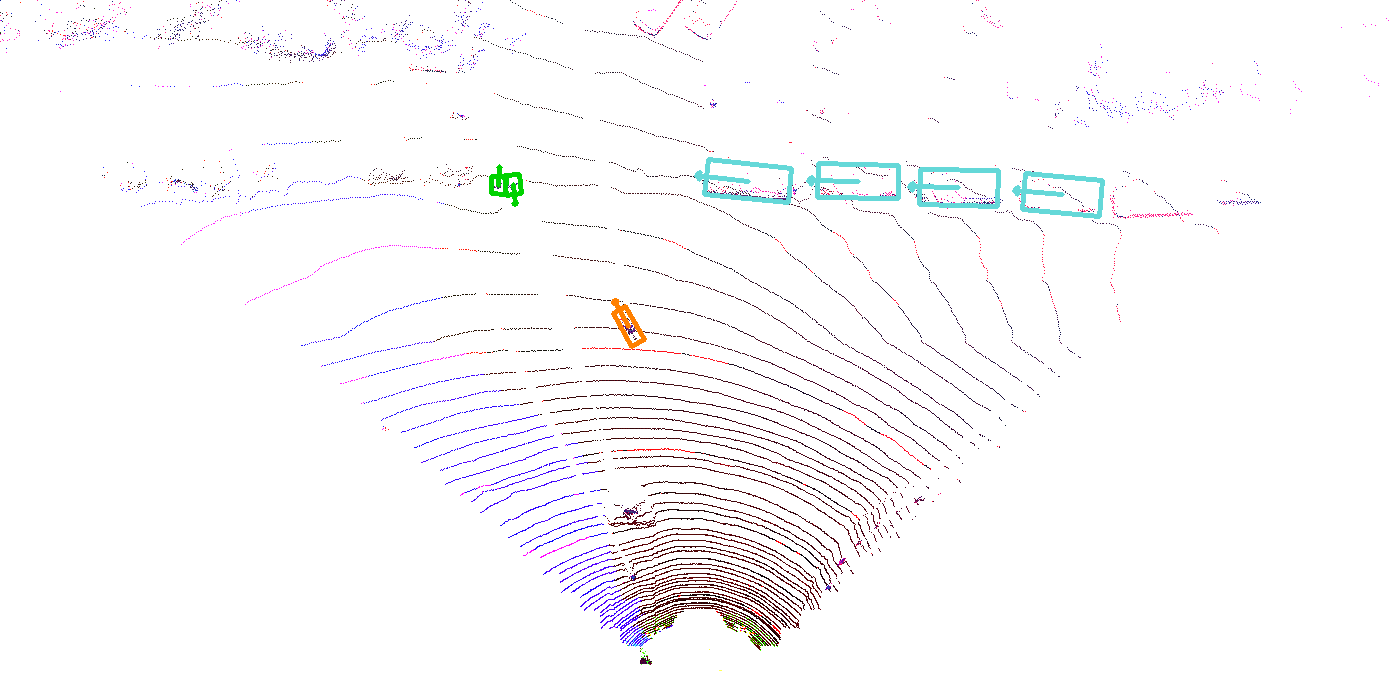}
\label{subfig:testa}}
\subfloat[]{%
\includegraphics[width=0.328\linewidth]{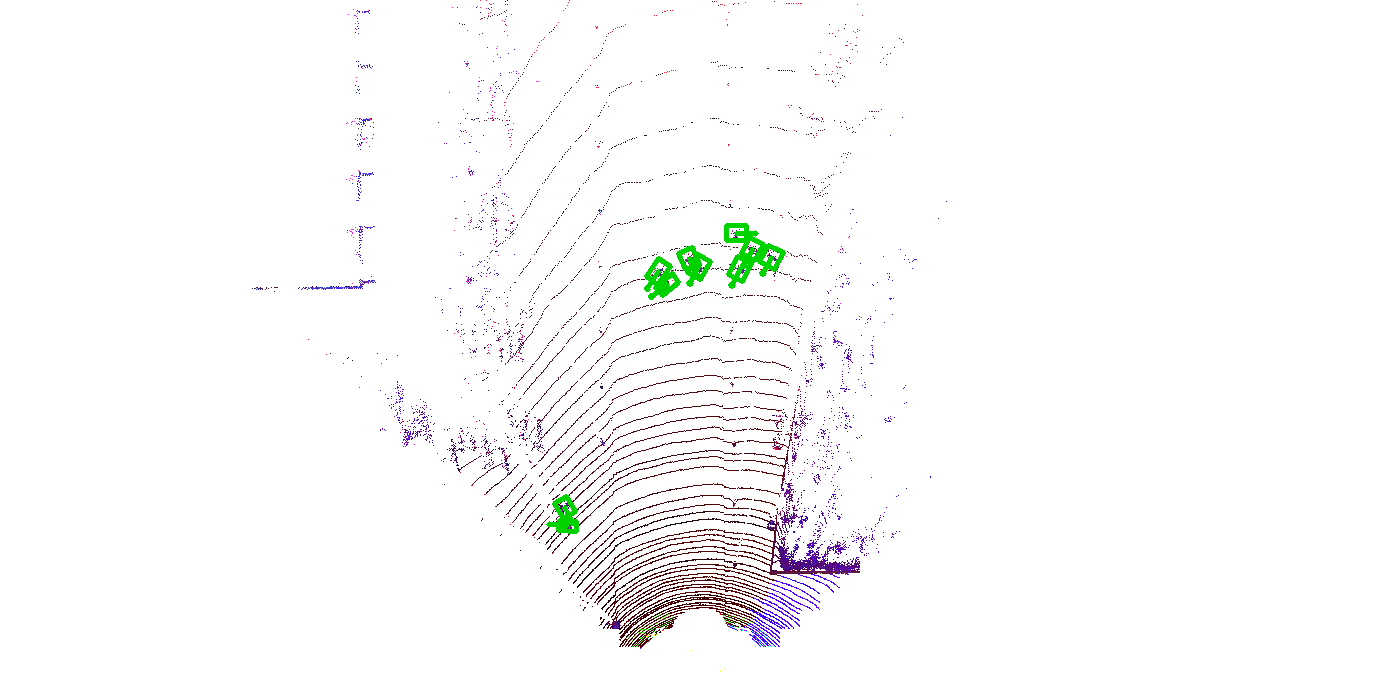}
\label{subfig:testb}}
\subfloat[]{%
\includegraphics[width=0.328\linewidth]{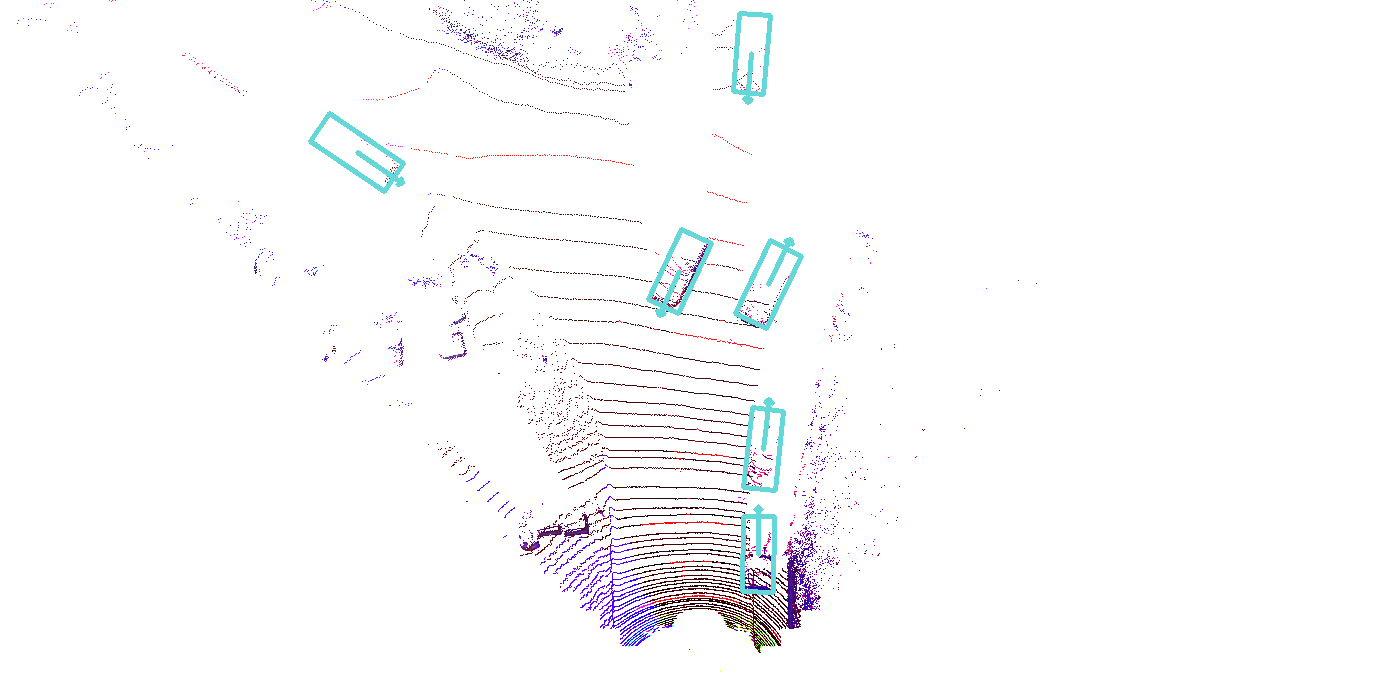}
\label{subfig:testc}}
\\
\subfloat{
\\
\includegraphics[width=0.328\linewidth]{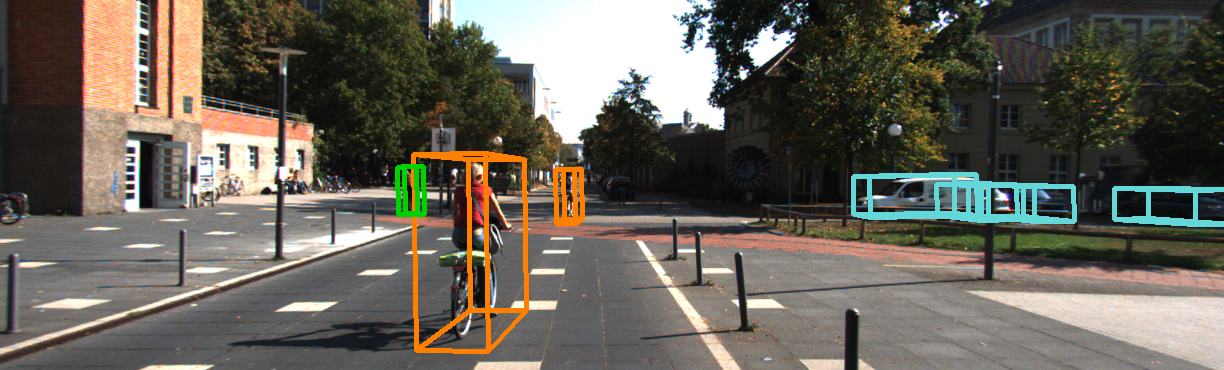}
\includegraphics[width=0.328\linewidth]{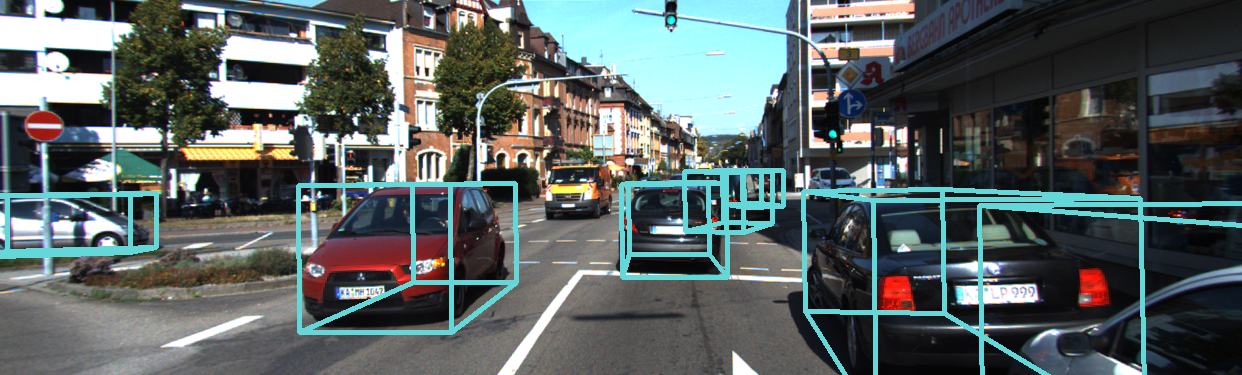}
\includegraphics[width=0.328\linewidth]{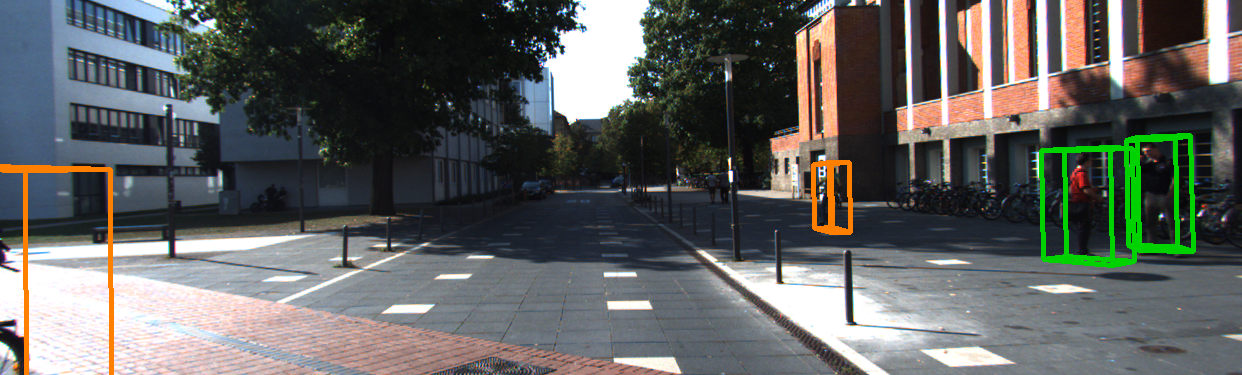}
}
\setcounter{subfigure}{3}
\\
\subfloat[]{%
\includegraphics[width=0.328\linewidth]{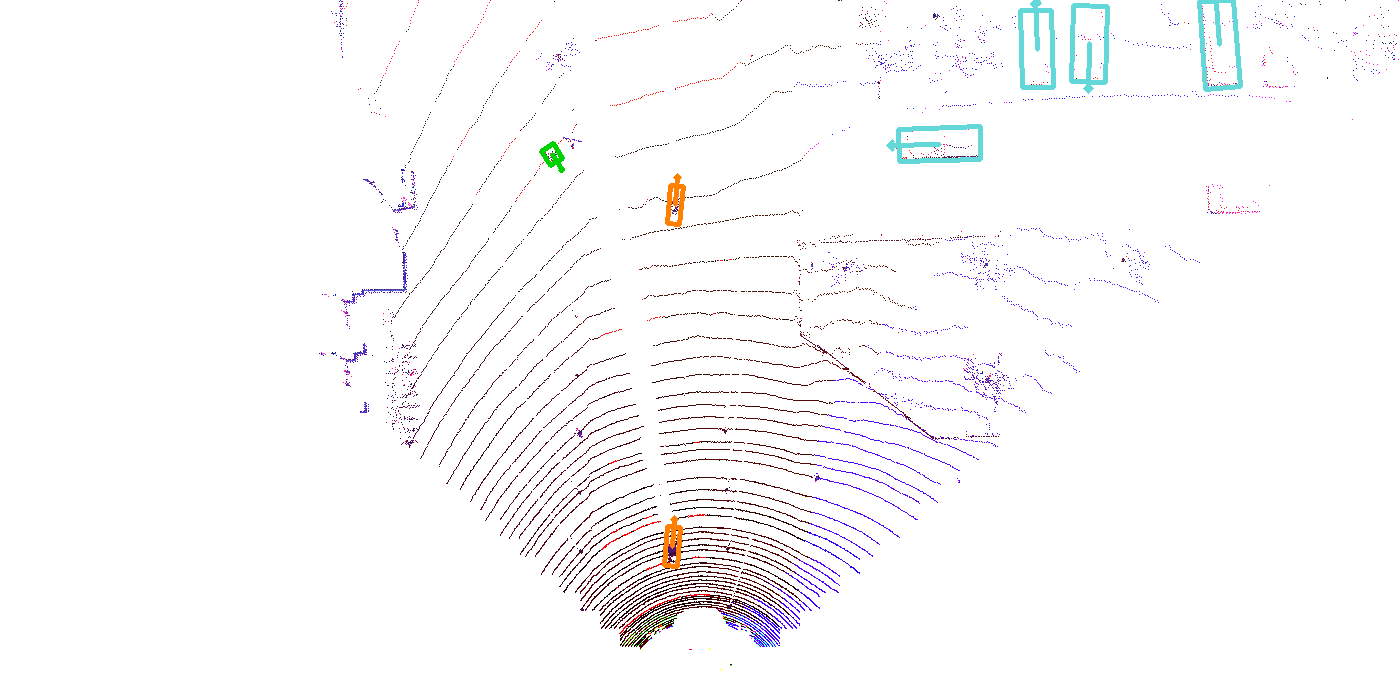}
\label{subfig:testd}}
\subfloat[]{%
\includegraphics[width=0.328\linewidth]{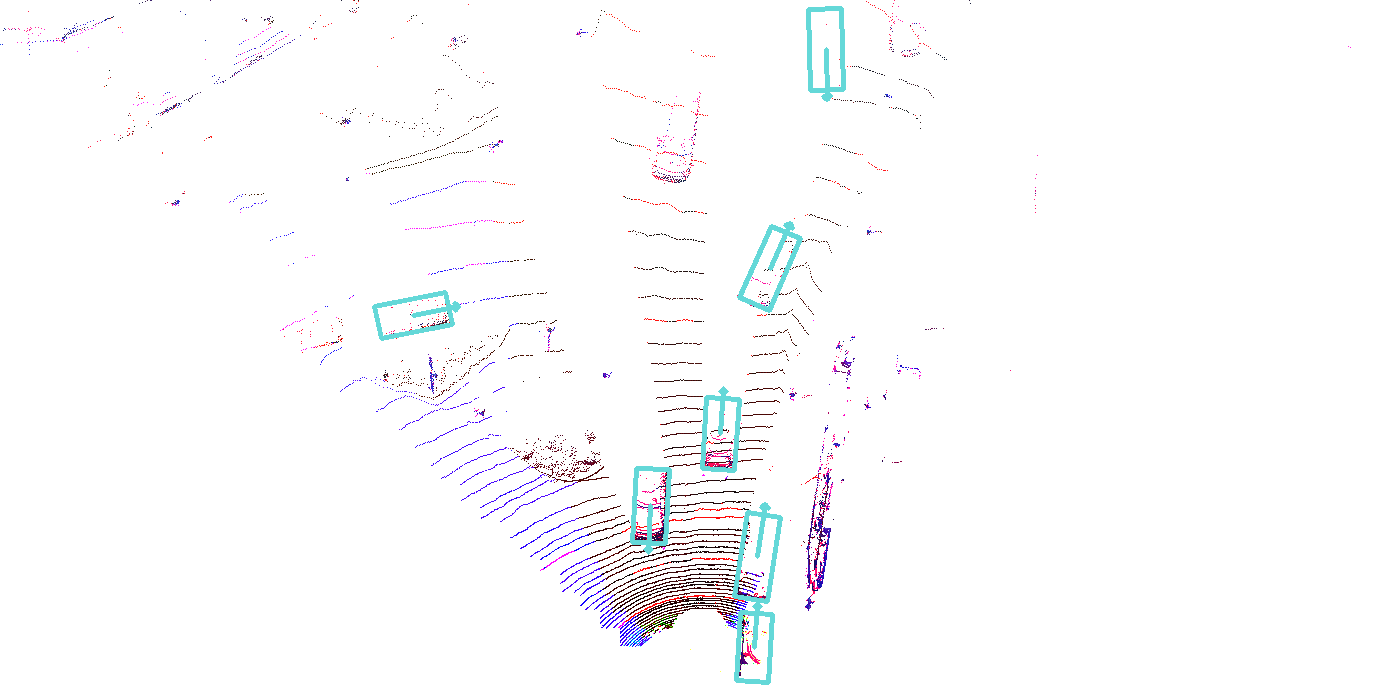}
\label{subfig:teste}}
\subfloat[]{%
\includegraphics[width=0.328\linewidth]{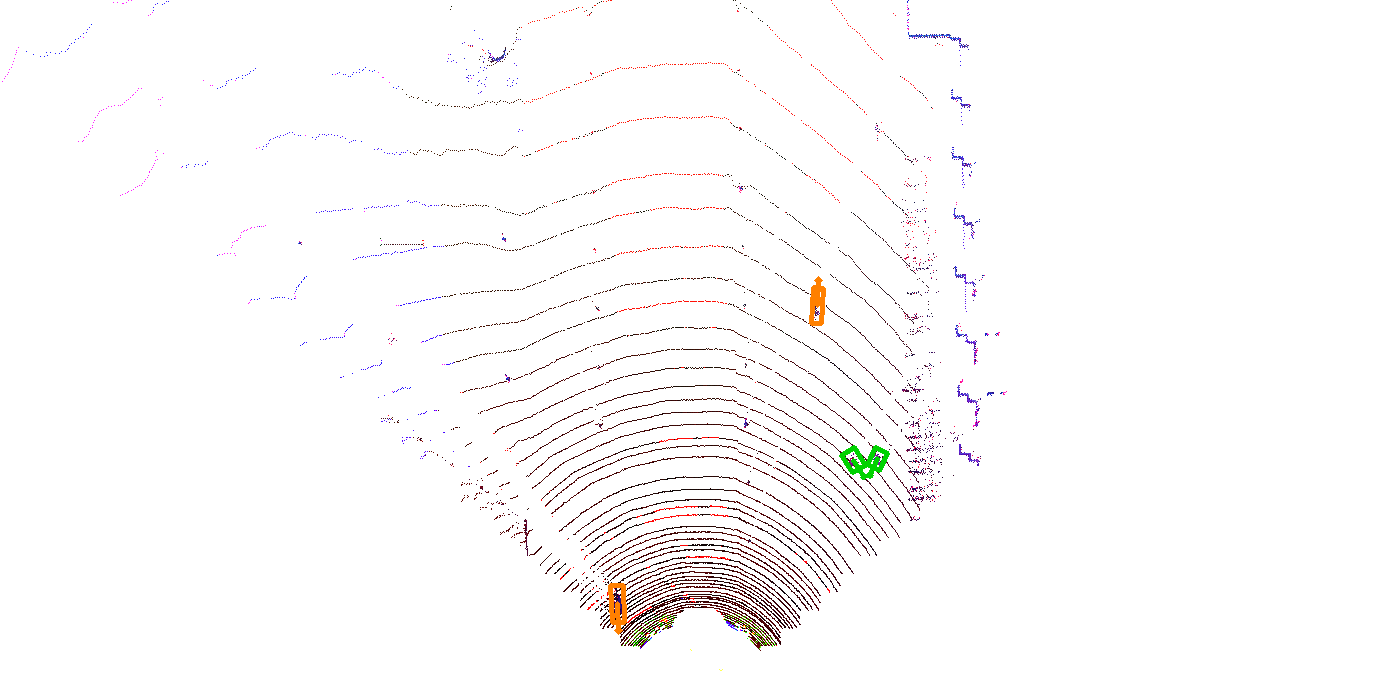}
\label{subfig:testf}}

\caption{Qualitative results of Birdnet+ on several scenes in KITTI test set. Both 3D detections (top) and BEV detections (bottom) are colored according to the class they belong to.}
\label{fig:testresults}
\end{figure*}

The validity of the proposed method is further confirmed by the qualitative results presented in Fig.~\ref{fig:testresults}, where 3D boxes are projected onto both the camera images (just for visualization purposes) and the range-limited BEV inputs. 

As it is based solely on LiDAR, our method does not get affected by the luminosity of the scene; instead, its performance depends largely on the number of points representing the objects. The high resolution featured by the BEV images allows detecting objects that are partially occluded or barely represented in the point cloud, as in Fig.~\ref{subfig:testc} and Fig.~\ref{subfig:teste}, and eases the segmentation between objects, as in Fig.~\ref{subfig:testf}. This projection also proves suitable to discriminate among different categories (e.g., cars and vans). 

On the other hand, the detection capabilities of the approach are naturally limited to the range represented in the BEV. Thus, Fig.~\ref{subfig:testa} and Fig.~\ref{subfig:teste} show some misdetections of distant objects that appear truncated. Nonetheless, these examples illustrate the overall excellent performance of the proposed approach in the estimation of 3D  boxes, especially with regard to the challenging issue of height estimation.

\section{CONCLUSIONS}

In this paper, we have presented BirdNet+, an end-to-end 3D object detection network able to classify and locate cars, pedestrians, and cyclists using LiDAR information. 


To the best of our knowledge, this work is the first one to perform full 3D box regression using only BEV images as input. The presented method relies on a two-stage detection architecture where axis-aligned proposals from an RPN go through a subsequent refinement step to obtain the final 3D oriented boxes. 


Experimental results on the KITTI dataset confirm the soundness of this approach. The different parameters defining the objects' 3D boxes have been successfully embedded into the inference framework, and, as a result, BirdNet+ behaves significantly better than its precursor framework and obtains comparable performance to similar methods.

In future work, more detailed cell encodings will be explored to enhance the representation capabilities of the BEV structure. Eventually, features from the raw cloud will be used during the final 3D box regression stage so that the outcome benefits from the full geometry information captured by LiDAR sensors, thus overcoming the loss of information caused by cloud discretization.






\section*{ACKNOWLEDGMENT}
Research supported by the Spanish Government (TRA2016-78886-C3-1-R and RTI2018-096036-B-C21),  the Universidad Carlos III of Madrid (PEAVAUTO-CM-UC3M) and the Comunidad de Madrid (SEGVAUTO-4.0-CM P2018/EMT-4362). We gratefully acknowledge the support of NVIDIA Corporation with the donation of the GPUs used for this research.


\bibliographystyle{IEEEtran}
\bibliography{paper}

\end{document}